\documentclass{article}
\usepackage{kr}
\usepackage{comment}
\usepackage{times}
\usepackage{soul}
\usepackage{url}
\usepackage[hidelinks]{hyperref}
\usepackage[utf8]{inputenc}
\usepackage[small]{caption}
\usepackage{graphicx}
\usepackage{amsmath}
\usepackage{amsthm}
\usepackage{booktabs}
\usepackage{algorithm}
\usepackage{algorithmic}
\usepackage{xcolor}

\newtheorem{example}{Example}
\newtheorem{theorem}{Theorem}
\newtheorem{definition}{Definition}
\newtheorem{proposition}{Proposition}
\newtheorem{corollary}{Corollary}
\newtheorem{principle}{Principle}
\newtheorem{lemma}{Lemma}

\usepackage{tikz}
\usetikzlibrary{positioning, calc}
\usepackage{enumitem}
\usepackage{amssymb}   

\title{Subargument Argumentation Frameworks: Separating Direct Conflict from Structural Dependency}

\author{%
Beishui Liao\\ 
Zhejiang University\\
}

\begin{document}

\maketitle

\begin{abstract}

Dung’s abstract argumentation frameworks model acceptability solely in terms of an attack relation, thereby conflating two conceptually distinct aspects of argumentative reasoning: direct conflict between arguments and the structural dependencies that arise from their internal composition. While this abstraction preserves extension-based semantics, it obscures how justification is grounded in subarguments and how defeats propagate through argument structure. We introduce Subargument Argumentation Frameworks (SAFs), an abstract framework in which direct attack and subargumenthood are represented as independent primitive relations. This separation makes structural dependency explicit at the representational level while leaving its semantic impact to be determined by structure-sensitive notions of defence, admissibility, and complete semantics defined within the framework. We show that projecting SAFs onto attack-only frameworks yields extension-equivalent Dung frameworks under all standard semantics, yet
the projection irreversibly loses information about justificatory grounding and structural propagation. SAFs therefore provide strictly greater representational expressiveness while remaining semantically compatible with Dung’s theory, thereby offering a principled basis for
structure-sensitive accounts of defence, justification, and explanation in abstract argumentation.
\end{abstract}

\section{Introduction}
Abstract argumentation has become the standard abstraction for nonmonotonic reasoning with arguments because it separates argument evaluation from the internal structure of arguments. By representing arguments as atomic entities connected only by an attack relation, it provides a uniform semantic foundation applicable across domains, logics, and application contexts~\cite{Dung1995}. This deliberate abstraction, however, fixes the signature of abstract argumentation to a single relational dimension, excluding any independent representation of structural dependency. Consequently, information about how arguments are constructed and how conflicts interact with that structure is absent at the abstract level. In structured argumentation formalisms such as ASPIC$^+$ and DeLP~\cite{ModgilPrakken2014,GarciaSimari2004}, this information is reintroduced indirectly: subarguments determine when attacks are induced, propagated, or lifted to larger arguments, and notions such as rebuttal or undercut rely implicitly on internal structure. Structural dependencies therefore affect acceptability only through derived attack relations. While effective in concrete instantiations, this strategy conflates conflict with structure at the abstract level, leaving no independent dimension for representing structural dependency and making it difficult to determine whether a semantic outcome arises from genuine conflict or merely structural propagation.

This paper starts from the observation that extension-based equivalence \cite{DBLP:journals/ai/BaumannDLW19} does not imply structural or explanatory equivalence. Different subargument configurations may collapse into the same Dung framework, yielding identical extensions while differing in how attacks are grounded, how defence propagates, and how acceptance is justified. These differences become crucial in settings concerned with explainability, modularity, and dynamic updates, where understanding \emph{why} an argument is accepted matters as much as \emph{whether}. To address this limitation, we adopt a different abstraction stance. Rather than encoding structure into attacks or committing to a particular logical instantiation, we extend the abstract signature by treating subargument relations as a primitive component of abstract argumentation orthogonal to conflict. This yields a minimal two-relational extension of Dung’s framework in which direct attacks represent explicit conflict and a subargument relation represents structural dependency. Crucially, attacks on subarguments are not assumed to induce attacks on superarguments at the representation level; such effects are captured semantically, making conflict and structural dependency independent modelling dimensions.

Building on this separation, we develop structure-sensitive notions of
defence, admissibility, and fixpoint semantics that preserve the core
architecture of Dung’s theory.
Defence is grounded in structurally relevant subarguments rather than inferred attack
patterns, and standard semantic guarantees, including monotonicity and the
existence of grounded extensions, are retained.
At the same time, we show that structural information cannot be recovered once
it is forgotten via a forgetful projection: although subargument relations can be erased without
affecting extensions, they cannot be encoded back into attack-only frameworks
without loss.
This establishes not merely a technical separation result, but a strict
increase in representational expressiveness obtained by extending the abstract
signature with an independent structural dimension.

\paragraph*{Contributions.} This paper makes four main contributions:
\begin{enumerate}
    \item It introduces \textbf{Subargument Argumentation Frameworks (SAFs)}, an abstract model that explicitly represents subargument relations alongside direct attacks. SAF is not a semi-structured framework, but a structure-aware abstract one that extends the signature of abstract argumentation.
    \item It develops a \textbf{structure-sensitive semantics} that respects subargument commitment while preserving Dung's fixpoint characterisation, thereby yielding a conservative semantic extension.
    \item It formally analyses the effect of \textbf{forgetting structure}, proving that extensions are preserved but information about grounding, propagation, and justification is irrecoverably lost, demonstrating a strict gain in expressive power induced by the additional structural dimension.
    \item It provides a \textbf{principled explanatory perspective}, showing how structural dependency can be separated from conflict handling and why introducing an independent structural axis is essential for explanation-aware reasoning.
\end{enumerate}

The paper is structured as follows. Section~\ref{sec:preliminaries} recalls basic notions. Section~3 presents a motivating example. Section~4 introduces SAFs and their semantics. Section~5 analyses expressiveness. Section~6 offers a principles-based interpretation. Section~7 discusses related work, and Section~8 concludes.


\section{Preliminaries}
\label{sec:preliminaries}

We recall the basic notions of Dung's abstract argumentation, which serves as the semantic baseline throughout the paper~\cite{Dung1995}. We also fix notation for closure operators used later.

\begin{definition}[AAF]
An \emph{abstract argumentation framework} (AAF) is a pair $\mathcal{G}=(A, Att)$, where $A$ is a finite set of arguments and $Att \subseteq A \times A$ is a binary attack relation.
\end{definition}

\begin{definition}[Conflict-freeness and Defence]
A set $E \subseteq A$ is \emph{conflict-free} if  there are no $ a,b \in E$ such that $(a,b) \in Att$. An argument $a \in A$ is \emph{defended} by $E$ if for every $b \in A$ with $(b,a) \in Att$, there exists $c \in E$ such that $(c,b) \in Att$.
\end{definition}

\begin{definition}[Admissible and Complete Extensions]
A conflict-free set $E \subseteq A$ is \emph{admissible} if it defends all its elements. An admissible set $E$ is a \emph{complete extension} if it contains every argument it defends. 
\end{definition}

\begin{definition}[Grounded, Preferred and Stable Extensions]
The \emph{grounded extension} is the $\subseteq$-minimal complete extension,
and a \emph{preferred extension} is a $\subseteq$-maximal complete extension.
A conflict-free set $E \subseteq A$ is a \emph{stable extension} if it attacks
every argument in $A \setminus E$.
\end{definition}

Let $\sigma_\mathsf{AAF}(\mathcal{G})$ denote the set of $\sigma_\mathsf{AAF}$-extensions of
$\mathcal{G}$, where $\sigma_\mathsf{AAF} \in \{\text{cmp}, \text{grd},
\text{prf}, \text{stb}\}$ denote complete, grounded,
preferred, and stable semantics, respectively.

The characteristic function $\mathsf{F}_{\mathcal{G}}: 2^{A} \to 2^{A}$ is defined as $\mathsf{F}_{\mathcal{G}}(E) = \{ a \in A \mid a \text{ is defended by } E \}$. It is monotone w.r.t. set inclusion, and $E$ is a complete extension iff $E = \mathsf{F}_{\mathcal{G}}(E)$. Its least fixpoint is the grounded extension.

Given a binary relation $R \subseteq A \times A$, we denote its reflexive and transitive closure by $R^{*}$. For $x \in A$, $\mathrm{Reach}_R(x) = \{ y \in A \mid (x, y) \in R^{*} \}$ denotes the set of nodes reachable from $x$ via $R^{*}$.


\section{A Motivating Example}
\label{sec:motivating}

We illustrate why compiling subargument structure into attacks fails to
capture conflict grounding and propagation, even when extension-based
outcomes coincide.

Consider arguments $\{a, b_1, b_2, b_3, c\}$ with a proper subargument
relation such that $b_1$ is a direct subargument of $b_2$, and $b_2$ and
$c$ are direct subarguments of $b_3$.
Assume two direct attacks: $a$ attacks $b_1$, and $b_3$ attacks $a$, with
no other attacks present. Here, 
only direct (irreducible) attacks targeting minimal subarguments are
considered.
The resulting configuration, called an SAF, is shown in Fig.~\ref{fig:example}(a).

\begin{figure}[t]
\centering
\begin{tikzpicture}[
    node distance=0.6cm,
    arg/.style={draw, circle, minimum size=0.5cm, font=\small},
    attack/.style={->, thick, red, shorten >=1pt, shorten <=1pt},
    subarg/.style={->, dashed, blue, shorten >=1pt, shorten <=1pt}
]
\begin{scope}[local bounding box=left]
    \node[arg] (b1) {$b_1$};
    \node[arg, below=of b1] (b2) {$b_2$};
    \node[arg, below=of b2] (b3) {$b_3$};
    \node[arg, left=of b2] (a) {$a$};

    \node[arg, right=of b3] (c) {$c$};

    \draw[attack] (a) -- (b1);
    \draw[attack] (b3) -- (a);
    \draw[subarg] (b1) -- (b2);
    \draw[subarg] (b2) -- (b3);

    \draw[subarg] (c) -- (b3);

    \node[below=0.1cm of b3, font=\small] {(a) SAF};
\end{scope}

\begin{scope}[xshift=3cm, local bounding box=right]
    \node[arg] (b1r) {$b_1$};
    \node[arg, below=of b1r] (b2r) {$b_2$};
    \node[arg, below=of b2r] (b3r) {$b_3$};
    \node[arg, left=of b2r] (a2) {$a$};

    \node[arg, right=of b3r] (cr) {$c$};

    \draw[attack] (a2) -- (b1r);
    \draw[attack] (a2) -- (b2r);
    \draw[attack] (a2) -- (b3r);
    \draw[attack] (b3r) -- (a2);

    \node[below=0.1cm of b3r, font=\small] {(b) AAF};
\end{scope}

\end{tikzpicture}
\caption{\label{fig:example} (a) A framework with an explicit subargument relation and direct attacks. (b) The corresponding AAF obtained by propagating attacks along subargument relations.}
\end{figure}


If the subargument relations are forgotten, structural dependencies can be simulated by propagating attacks upward along the subargument hierarchy. That is, whenever an argument is attacked, all of its superarguments are
treated as attacked as well. Under this abstraction principle, the SAF in Fig.~\ref{fig:example}(a) induces
the AAF shown in Fig.~\ref{fig:example}(b), with attack set
$\{(a, b_1), (a, b_2), (a, b_3), (b_3, a)\}$.
Both frameworks yield the same complete extension $\{a\}$ under Dung semantics. However, despite this extension-level agreement, the two representations differ
crucially in three respects:

\begin{enumerate}[topsep=2pt,itemsep=0pt,parsep=0pt]
    \item \textbf{Grounding of conflict.} In the SAF, the conflict originates at $b_1$. The attack on $b_3$ is only indirect, arising from its dependency on $b_1$. In the AAF, all $b_i$ appear as equally direct targets of $a$, obscuring the actual point of conflict.

    \item \textbf{Defence and commitment.} In the SAF, accepting $b_3$ would require its entire subargument basis, including $b_1$  (via $b_2$) as well as the additional subargument $c$. In the AAF, $b_3$ is defended by its counterattack on $a$, regardless of the unresolved attack on $b_1$. Hence, defence is detached from structural commitment.

    \item \textbf{Propagation of status changes.} In the SAF, if $b_1$ were defeated (e.g., by another newly added attacker), the defeat would propagate through the subargument structure to all superarguments that structurally depend on the defeated component. In the AAF, no such dependency exists; the status of $b_2$ and $b_3$ is independent of $b_1$ once attacks are compiled.
\end{enumerate}

These distinctions are invisible at the level of extensions, yet they matter for \emph{explanation} (why is an argument accepted?), \emph{modularity} (how are arguments built from components?), and \emph{dynamics} (how does a local change affect the whole structure?). The example shows that attack‑only abstraction conflates conflict with structure, losing information that is essential for these purposes. 

To recover this information without committing to a specific structured formalism, the next section introduces SAFs, where a subargument relation is represented explicitly alongside attacks, and semantics are defined to respect structural dependencies.

\section{Subargument Argumentation Frameworks and Structure-sensitive Semantics}
\label{sec:saf}
Addressing the limitation identified in Section~3 calls for a representational refinement rather than a change of semantics. We therefore introduce an abstract argumentation framework that makes structural dependency explicit while preserving the core abstraction of
Dung’s theory. The central idea is to separate two orthogonal dimensions of argumentation: \emph{conflict}, represented by direct attacks, and \emph{structure},
represented by a subargument relation.
By treating subargumenthood as a primitive component, structural information is retained at the abstract level rather than encoded indirectly via attack patterns.
The proposed framework addresses three representational deficiencies identified in Section~3: (i) the absence of defence grounded in subarguments, (ii) the lack of structural propagation beyond direct conflict, and (iii) the collapse of structurally distinct configurations under attack-only abstraction.

\subsection{Subargument Argumentation Framework}
In the present framework, the subargument relation $\mathit{Sub}$ is treated as
a primitive capturing only direct structural dependencies.
It is therefore neither reflexive nor transitive.
Structural closure is represented separately by its reflexive-transitive
closure $\mathit{Sub}^*$ and is invoked only at the semantic level.
Accordingly, all notions of propagation, grounding, and justification are
understood with respect to $\mathit{Sub}^*$ rather than the primitive relation
itself.

This modelling choice contrasts with structured formalisms such as ASPIC$^+$,
where subargument relations are reflexive and transitive by construction.
Our aim, however, is not to mirror instance-level syntax, but to isolate
structural dependency as an abstract representational component orthogonal to
conflict.
Keeping $\mathit{Sub}$ primitive separates atomic structural inputs from their
derived closure effects, which are reconstructed via $\mathit{Sub}^*$.

This distinction becomes essential once updates are considered.
A single local change to a direct dependency may induce many consequences in
the transitive closure.
By keeping $\mathit{Sub}$ minimal and invoking $\mathit{Sub}^*$ only
semantically, the framework preserves a clear notion of where structural change
originates, which is crucial for analysing propagation and justification under
abstraction.

Throughout the paper, we distinguish primitive, direct attacks $D$ from induced
attacks.
While $D$ itself is not closed under subargument relations, the generic attack
relation $\mathit{Att}$ in standard Dung frameworks may already encode
structure-induced interactions.

\begin{definition}[SAF]
\label{def:saf}
A \emph{Subargument Argumentation Framework} (SAF) is a triple
$\mathcal{F}=(A,D,\mathit{Sub})$, where:
\begin{itemize}
  \item $A$ is a finite set of arguments;
  \item $D \subseteq A \times A$ is a direct attack relation;
  \item $\mathit{Sub} \subseteq A \times A$ is a direct and proper subargument relation,
  assumed to be irreflexive and non-transitive.
\end{itemize}
The relations $D$ and $\mathit{Sub}$ are independent primitives.
The reflexive and transitive closure of $\mathit{Sub}$ is denoted $\mathit{Sub}^{*}$.

\end{definition}

By separating conflict and structure at the representational level, the
framework preserves the information required for reasoning about defence and
commitment, and provides the basis for the structure-sensitive notions of
defence and admissibility developed next.

The representational choices underlying SAFs are not ad hoc.
Both rule-based (e.g., ASPIC$^+$) and assumption-based (e.g., ABA) paradigms already exhibit primitive structural dependencies and irreducible attack points. SAFs make this separation explicit by recording only direct attacks in $D$ and direct proper subarguments in $\mathit{Sub}$, leaving structural propagation to the semantic level rather than compiling it into the attack relation. (Detailed illustration is provided in the \emph{supplementary material})

\subsection{Structure-sensitive Defense}
The framework introduced in Section 4.1 makes structural
dependencies between arguments explicit. However, as illustrated by the motivating example in Section 3, representing structure alone is not sufficient: the central difficulty lies in how defence and acceptability should take structural dependencies into account.

In particular, Section 3 shows that counterattacking an attacker at the surface level does not necessarily resolve a conflict that is grounded in an attacked subargument. This observation motivates the need for a notion of defence that is sensitive to subargument structure, while remaining compatible with the core semantic architecture of Dung’s theory.

We first address the defence-grounding problem identified
in Section 3 by refining the notion of defence so that attacks
on subarguments cannot be neutralised without being structurally addressed. The reflexivity of $\mathit{Sub}^*$ allows us to uniformly treat attacks
directed at an argument itself and attacks directed at its subarguments, thereby subsuming Dung-style defence as a special case.

\begin{definition}[Structure-sensitive defence]\label{def:ss-defense}
Let $\mathcal{F} = (A, D, \mathit{Sub})$ be an SAF and $E\subseteq A$.
An argument $a\in A$ is \emph{defended by $E$} iff
for every $x\in \mathit{Sub}^*(a)$ and every $b\in A$ such that $(b,x)\in D$,
there exist $c\in E$ and $b'\in \mathit{Sub}^*(b)$ with $(c,b')\in D$.
\end{definition}


This definition refines Dung’s notion of defence by making it sensitive to
attacks directed at an argument’s subargument structure, rather than only to
attacks on the argument itself.
Defence remains an external notion, and does not require the defended argument
to belong to the defending set.
Instead, it exposes the structural grounds on which justificatory support is
established at the abstract level, while remaining compatible with
extension-based semantics.

\begin{example}[Structure-sensitive defence in Fig.~1] \label{ex:struct-conflict}
Consider the SAF shown in Fig.~1.
Although $b_3$ is not directly attacked, its defence is not automatic.
According to Definition~\ref{def:ss-defense}, defending $b_3$ requires that
every attack on any $x \in \mathit{Sub}^*(b_3)$ be countered.

Since $(a,b_1)\in D$ and $b_1\in \mathit{Sub}^*(b_3)$, defence requires a
counterattack against $a$ or one of its subarguments.
This condition is satisfied in Fig.~1, as $b_3$ directly attacks $a$.
Hence, the attack on the subargument $b_1$ is successfully countered, and $b_3$
is defended.

This example shows that structure-sensitive defence takes attacks on
subarguments into account, and that counterattacks may originate from
superarguments rather than from the attacked subarguments themselves.
\end{example}

\subsection{Structure-sensitive Admissibility and the Characteristic Function}

We next revisit conflict-freeness and admissibility in the
presence of subargument relations. Since accepting an argument implicitly commits one to accepting its subarguments,
conflicts must be evaluated relative to the full structural closure of arguments. 

The resulting notion of admissibility internalises a commitment to subarguments: accepting an argument entails accepting its structural basis.

\begin{definition}[Structure-sensitive conflict-freeness]
\label{def:ss-cf}
Let $\mathcal{F} = (A, D, \mathit{Sub})$. Two arguments $a,b \in A$ are \emph{in conflict} iff $\exists x \in \mathit{Sub}^{*}(a), \exists y \in \mathit{Sub}^{*}(b)$ such that  $(x,y) \in D$.  
$E \subseteq A$ is \emph{conflict-free} if no two members of $E$ are in conflict.
\end{definition}

This notion ensures that conflicts arising at any structural
level are taken into account, even when attacks are defined
only as direct conflicts.

\begin{definition}[Admissibility]\label{def:ss-adm}
A set $E\subseteq A$ is \emph{admissible} iff
\begin{enumerate}
  \item $E$ is conflict-free;
  \item every $a\in E$ is defended by $E$;
  \item $E$ is closed under $\mathit{Sub}^*$, i.e. $a\in E \Rightarrow \mathit{Sub}^*(a)\subseteq E$.
\end{enumerate}
\end{definition}

Under this definition, admissibility internalises subargument commitment:
accepting an argument entails accepting its structural basis. This commitment
is reflected not only in conflict detection, as illustrated in
Example~\ref{ex:struct-conflict}, but also in the requirement that accepted
arguments be defended in a manner that accounts for their subargument structure.

Importantly, subargument closure is imposed here as a semantic requirement on admissible sets, rather than as an independent syntactic constraint. This design ensures that admissibility enforces structural coherence among accepted arguments, while leaving the operational behaviour of defence to the underlying characteristic function.

Despite this refinement, the overall fixpoint-based architecture of Dung’s
semantics is preserved.

\begin{definition}[Characteristic function]
\label{def:char-fn}
Let $\mathsf{Def}_{\mathcal{F}}(E) = \{ a \in A \mid a \text{ is defended by } E \}$.  
The \emph{characteristic function} of $\mathcal{F}$ is $\mathsf{F}_{\mathcal{F}}: 2^{A} \to 2^{A}$ defined by $\mathsf{F}_{\mathcal{F}}(E) = \mathsf{Def}_{\mathcal{F}}(E)$.
\end{definition}

It is crucial to distinguish the role of subargument closure at different
levels of the framework. In Definition~\ref{def:ss-adm}, subargument closure
functions as a normative constraint on admissible sets, ensuring that any
set claiming acceptability also commits to all of its subarguments.
By contrast, the characteristic function captures this commitment from an
operational perspective. The following proposition shows that
structure-sensitive defence preserves subargument closure already at the
level of the characteristic operator itself.

\begin{proposition}\label{prop:subarg-closure}
Let $\mathcal{F} = (A,D,\mathit{Sub})$ be an SAF and $E \subseteq A$.
Then the operator $\mathsf{F}_{\mathcal{F}}(E) = \mathrm{Def}_{\mathcal{F}}(E)$ is
subargument-closed, that is,
\[
a \in \mathrm{Def}_{\mathcal{F}}(E)
\;\Rightarrow\;
\mathit{Sub}^*(a) \subseteq \mathrm{Def}_{\mathcal{F}}(E).
\]
\end{proposition}

\begin{proof}
Let $a \in \mathrm{Def}_{\mathcal{F}}(E)$ and let $a' \in \mathit{Sub}^*(a)$.
We show that $a' \in \mathrm{Def}_{\mathcal{F}}(E)$.

Fix an arbitrary argument $b \in A$ such that $(b,a') \in D$.
Since $a' \in \mathit{Sub}^*(a)$, the argument $b$ attacks a subargument of $a$.
By the definition of structure-sensitive defence, the fact that $a$ is
defended by $E$ entails that every attack on any subargument of $a$ is
countered by $E$.

Hence, there exists an argument $c \in E$ and an argument $b' \in
\mathit{Sub}^*(b)$ such that $(c,b') \in D$.
This exactly satisfies the defence condition for $a'$ with respect to $E$.

Since $b$ was chosen arbitrarily, it follows that $a'$ is defended by $E$.
Therefore, $a' \in \mathrm{Def}_{\mathcal{F}}(E)$, and consequently
$\mathit{Sub}^*(a) \subseteq \mathrm{Def}_{\mathcal{F}}(E)$.
\end{proof}





\begin{proposition}[Monotonicity]
\label{prop:monotone}
The operator $\mathsf{F}_{\mathcal{F}}$ is monotone with respect to set inclusion.
\end{proposition}

\begin{proof}
If $E\subseteq E'$, any counterattack witnessing the defence of an argument
by $E$ is also available from $E'$, which yields
$\mathsf{F}_{\mathcal{F}}(E)\subseteq \mathsf{F}_{\mathcal{F}}(E')$.
\end{proof}

\subsection{Complete Semantics and the Fixed-point Theorem}

\begin{definition}[Complete extension]
Let $F=(A,D,\mathit{Sub})$ be an SAF.
An admissible set $E\subseteq A$ is \emph{complete} iff
 $\forall a\in A$,
if $a$ is structure-sensitively defended by $E$, then $a\in E$.
\end{definition}

\begin{theorem}[Fixed-point characterisation]
\label{thm:fixed-point}
Let $\mathcal{F}$ be an SAF and $E \subseteq A$. The following are equivalent:
\begin{enumerate}
    \item $E$ is a complete extension of $\mathcal{F}$.
    \item $E$ is conflict-free and $E = \mathsf{F}_{\mathcal{F}}(E)$.
    \item $E$ is a conflict-free fixed point of $\mathsf{F}_{\mathcal{F}}$.
\end{enumerate}
\end{theorem}

\begin{proof}[Proof sketch]
The proof follows the standard argument for Dung’s fixpoint characterisation,
adapted to the structure-sensitive defence operator of the SAF
$\mathcal{F} = (A, D, \mathit{Sub})$.

($1 \Rightarrow 2$)
Let $E$ be a complete extension.
By Definition~8, $E$ is conflict-free and defends all its elements.
By Definition~10, completeness requires that every argument defended by $E$
belongs to $E$, hence $E = \mathsf{F}_{\mathcal{F}}(E)$.

($2 \Rightarrow 3$)
Immediate by definition of fixed points.

($3 \Rightarrow 1$)
Let $E$ be a conflict-free fixed point of $\mathsf{F}_{\mathcal{F}}$.
By $E = \mathsf{F}_{\mathcal{F}}(E)$, every argument in $E$ is defended by $E$.
Moreover, by Proposition~1, $\mathsf{F}_{\mathcal{F}}(E)$ is closed under
$\mathit{Sub}^*$, so $E$ satisfies the subargument-closure condition of
Definition~8.
Hence $E$ is admissible and contains all arguments it defends, and is therefore
a complete extension. (A full formal proof is provided in the \emph{supplementary material})
\end{proof}

\begin{corollary}
Every SAF $\mathcal{F}$ admits a grounded extension.
\end{corollary}
\begin{proof}
This follows immediately from Theorem~1 and the monotonicity of
$\mathsf{F}_{\mathcal{F}}$.
\end{proof}

This section shows that explicit structural information can be incorporated
into abstract argumentation by minimally extending Dung’s framework,
yielding a \emph{structure-sensitive semantics} while preserving its
fixed-point characterisation.

Although this semantics resolves the grounding and propagation issues
identified earlier, a natural question arises: can these structural distinctions
be recovered once structure is abstracted away? The next section shows that
they cannot.

Complete semantics serves as the basis for other standard Dung-style
semantics in SAFs. Preferred and stable semantics are defined in the usual
way from admissible sets, using the structure-sensitive notion of defence.

A \emph{preferred extension} is a $\subseteq$-maximal complete extension.
An admissible set $E$ is \emph{stable} iff  $\forall a \in A \setminus E$,
$\exists b \in E$ such that
$b$ structure-sensitively attacks $a$, i.e.,
there exists 
$x \in \mathit{Sub}^*(a)$ with $(b,x) \in D$.

We write $\sigma_\mathsf{SAF}(\mathcal{F})$ for the set of extensions of
$\mathcal{F}$ under $\sigma_\mathsf{SAF}$, where $\sigma_\mathsf{SAF} \in
\{\mathsf{CMP}, \mathsf{GRD},
\mathsf{PRF}, \mathsf{STB}\}$ denotes, respectively,
the structure-sensitive complete, grounded,
preferred, and stable semantics.

\section{Forgetting Structure and Structural Expressiveness}
\label{sec:expressiveness}

This section analyses the effect of forgetting structure and clarifies
what is lost when structural dependencies are collapsed into attack-only
representations. While subargument relations can be abstracted away
without affecting extension-based semantics, this preservation incurs a
representational cost. Distinct structural frameworks may induce the same
Dung framework and thus identical extensions, yet differ in how defence is
grounded, how conflicts propagate, and how arguments are structurally
justified—distinctions that cannot be recovered from the attack relation
alone. We therefore conclude that attack-only abstraction is strictly less
expressive than frameworks that represent structure explicitly, reflecting
an inherent loss of information at the abstract level.

\subsection{Forgetting the Subargument Relation}

We begin by formalising the process of forgetting subargument structure.
Importantly, forgetting is not understood here as a mere deletion of the
subargument relation. Instead, it is realised via a projection that compiles
the effects of structural dependency into the attack relation, yielding a
standard Dung framework.

\begin{definition}[Forgetful Projection]
\label{def:forgetful-projection}
Let \(\mathcal{F} = (A, D, \mathit{Sub})\). The \emph{forgetful projection} of \(\mathcal{F}\) is an AAF \(\Gamma(\mathcal{F}) = (A, D^\Gamma)\), where
\[
D^\Gamma = \{(a, b) \mid \exists b' \in \mathit{Sub}^{*}(b) \text{ s. t. } (a, b') \in D\}.
\]
\end{definition}

The projection $\Gamma(F)$ captures the cumulative effect of direct attacks
along subargument structure, while discarding information about the specific
structural levels at which these attacks originate.

We first establish that forgetting structure does not affect extensions.

\begin{proposition}[Preservation of extensions under structural forgetting]
\label{prop:forgetting-preserves-extensions}
Let $\mathcal{F}=(A,D,\mathit{Sub})$ be an SAF and $\Gamma(\mathcal{F})= (A,D^{\Gamma})$ be its forgetful projection.
It holds that
$\sigma_\mathsf{SAF}(\mathcal{F})=\sigma_\mathsf{AAF}(\Gamma(\mathcal{F}))$.
\end{proposition}

\begin{proof}[Proof sketch]
The central observation is that structure-sensitive defence in
$\mathcal{F}$ coincides with standard Dung defence after projection.

For any set $E \subseteq A$ and any argument $a \in A$, we show that $a$ is
defended by $E$ in $\mathcal{F}$ if and only if $a$ is defended by $E$ in
$\Gamma(\mathcal{F})$. By construction of $D^{\Gamma}$, every
structure-sensitive attack on $a$ in $\mathcal{F}$ corresponds to an
attack on $a$ in $\Gamma(\mathcal{F})$, and conversely. Hence, the sets of
defended arguments induced by $E$ coincide in the two frameworks. It follows
that the characteristic functions associated with $\mathcal{F}$ and
$\Gamma(\mathcal{F})$ yield the same result on every input set.

By the fixpoint characterisation of complete semantics, the two frameworks
therefore admit the same complete extensions. Since the grounded extension is
the least complete extension and preferred extensions are the maximal complete
extensions, grounded and preferred semantics are preserved as well. Finally,
by definition of $D^{\Gamma}$, an admissible set attacks every
argument outside the set in $\mathcal{F}$ if and only if it does so in
$\Gamma(\mathcal{F})$. Hence, stable extensions coincide in the two frameworks.
(A full formal proof is provided in the \emph{supplementary material})
\end{proof}



Proposition~\ref{prop:forgetting-preserves-extensions} establishes that
forgetting subargument structure via the projection $\Gamma$
preserves all standard extension-based semantics.
Thus, when attention is restricted to final extensions,
explicit subargument relations do not affect the resulting sets of accepted arguments.

This preservation, however, does not extend to admissibility as a notion of
justified acceptance.
Consider the SAF $\mathcal{F}=(A,D,\mathit{Sub})$ from Example~1 and
let $E=\{b_3\}$.
In the projected AAF $\Gamma(\mathcal{F})$, $E$ is admissible,
since $b_3$ counterattacks its only attacker $a$.
In contrast, $E$ is not admissible in $\mathcal{F}$ under Definition~8,
because admissibility in SAFs requires $\mathit{Sub}^*$-closure:
accepting $b_3$ commits the set to accepting all arguments in
$\mathit{Sub}^*(b_3)$ (including $b_1,b_2$ and $c$), which are not contained in $E$.

The crucial point is that forgetting structure preserves which arguments
can ultimately be accepted together, but not the structural commitments
that justify their acceptance.
In SAFs, admissibility internalises subargument commitment through
$\mathit{Sub}^*$-closure: accepting an argument entails accepting
its entire structural basis.
Such commitments cannot be expressed in attack-only abstractions.

\subsection{Expressiveness Beyond Extensions}

While forgetting preserves extensions, it does not preserve structural information. Crucially, distinct SAFs may collapse into the same AAF under the forgetful projection.

\begin{theorem}[Strict Increase in Structural Expressiveness]
\label{thm:strict-increase-expressiveness}
There exist two SAFs \(\mathcal{F}_1 = (A, D_1, \mathit{Sub}_1)\) and \(\mathcal{F}_2 = (A, D_2, \mathit{Sub}_2)\) such that \(\mathcal{F}_1 \neq \mathcal{F}_2\) and \(\Gamma(\mathcal{F}_1) = \Gamma(\mathcal{F}_2)\).
\end{theorem}

\begin{proof}
Consider $A=\{a,b,c\}$ and let $D=\{(b,a)\}$.
Define two SAFs $\mathcal{F}_1=(A,D,\mathit{Sub}_1)$ and
$\mathcal{F}_2=(A,D,\mathit{Sub}_2)$, where
$\mathit{Sub}_1=\varnothing$ and $\mathit{Sub}_2=\{(c,a)\}$.
Then $\mathcal{F}_1 \neq \mathcal{F}_2$, but
\[
\Gamma(\mathcal{F}_1) = (A,D) = \Gamma(\mathcal{F}_2).
\]
Thus, distinct structural configurations are collapsed under projection,
showing that subargument information cannot be recovered from
attack-only representations.
\end{proof}

Thus, \(\mathcal{F}_1\) and \(\mathcal{F}_2\) map to the same Dung AAF but are not equivalent at the structural level, as they induce different patterns of defense and structural commitment.

To formalize the informational loss, we define propagation reaches that depend on structure versus attacks.

\begin{definition}[Structural Propagation Reach]
\label{def:structural-reach}
Let \(\mathcal{F} = (A, D, \mathit{Sub})\) and \(x \in A\). The \emph{structural propagation reach} of \(x\) in \(\mathcal{F}\) is defined as
\[
\mathrm{Reach}_{\mathcal{F}}(x) = \{ y \in A \mid x \in \mathit{Sub}^{*}(y) \}.
\]
Intuitively, $\mathrm{Reach}_{\mathcal{F}}(x)$ contains exactly those arguments that structurally depend on $x$, i.e., the superarguments of $x$. It captures propagation induced purely by subargument
dependency, independently of conflict relations.
\end{definition}

\begin{definition}[Attack-based Propagation Reach]
\label{def:attack-reach}
Let \(\mathcal{G} = (A, Att)\) be an AAF and let \(x \in A\). The \emph{attack-based propagation reach} of \(x\) in \(\mathcal{G}\) is defined as
$
\mathrm{Reach}_{\mathcal{G}}(x) = \{ y \in A \mid \text{there exists a directed path from } x \text{ to } y \text{ in } Att \}.
$
\end{definition}

\begin{theorem}[Non-encodability of Structural Propagation]
\label{thm:non-encodability}
Let \(\Gamma\) be the forgetful projection from Definition~\ref{def:forgetful-projection}. There exists an SAF \(\mathcal{F} = (A, D, \mathit{Sub})\) and an argument \(x \in A\) such that
\[
\mathrm{Reach}_{\mathcal{F}}(x) \neq \mathrm{Reach}_{\Gamma(\mathcal{F})}(x).
\]
Hence structural propagation induced by subargument
dependencies is not preserved under the forgetful projection
and cannot, in general, be recovered from the attack
relation alone.
\end{theorem}

\begin{proof}
Consider the SAF in Fig.1. By Definition~\ref{def:structural-reach}, since \(b_2 \in \mathit{Sub}^{*}(b_3)\), we have \(\mathrm{Reach}_{\mathcal{F}}(b_2) = \{b_2, b_3\}\). Now consider the forgetful projection \(\Gamma(\mathcal{F})= (A, D^\Gamma)\). Its attack relation is \(D^\Gamma = \{(a, b_1), (a, b_2), (a, b_3), (b_3, a)\}\). In \(\Gamma(\mathcal{F})\), argument \(b_2\) does not participate in any attack, and there is no attack path starting from \(b_2\). Hence, by Definition~\ref{def:attack-reach}, \(\mathrm{Reach}_{\Gamma(\mathcal{F})}(b_2) = \{b_2\}\). Therefore, \(\mathrm{Reach}_{\mathcal{F}}(b_2) \neq \mathrm{Reach}_{\Gamma(\mathcal{F})}(b_2)\), demonstrating that   purely structural propagation
cannot be reconstructed from attack-based
propagation after forgetting subargument structure..
\end{proof}

\section{Principles and Explanatory Perspective}
\label{sec:principles-and-explanation}

Section~\ref{sec:expressiveness} showed that forgetting structure
preserves extension-based semantics while causing an invisible loss
at the extension level. To clarify what is lost and why this loss is
conceptually significant, we now adopt a principle-based perspective.

\subsection{Principles for Structure-Aware Argumentation}
\label{subsec:principles}

The preceding analysis established that explicitly representing
subargument structure increases expressive power, even when
final extensions coincide under abstraction.
We now characterise this gain in terms of structural principles.

These principles introduce no new normative requirements.
Rather, they articulate structural assumptions implicit in
structured reasoning that are violated by attack-only abstraction.
Each principle is satisfied by the structure-sensitive semantics
developed here, but fails once structure is forgotten.

\begin{principle}[Separation of Conflict and Structure]
\label{princ:separation}
An argumentation framework satisfies the \emph{Separation of Conflict and Structure} principle if conflict relations and structural-dependency relations are represented by independent primitives, and neither can be derived from the other.
\end{principle}

\begin{theorem}
\label{thm:principle-separation}
Every SAF $\mathcal{F}=(A,D,\mathit{Sub})$ satisfies Principle~\ref{princ:separation}.
\end{theorem}
\begin{proof}
By Definition~\ref{def:saf}, $D$ contains only direct attacks and is not closed under structural propagation. The subargument relation $\mathit{Sub}$ is introduced as a primitive and is not definable from $D$. Hence changes in $\mathit{Sub}$ do not affect $D$, and vice-versa, which establishes the independence of conflict and structure.
\end{proof}

\begin{principle}[Structural coverage of defence]
\label{princ:ss-defense}
Let $\mathcal{F}=(A,D,\mathit{Sub})$ be a SAF and let $E\subseteq A$.
If $a\in E$, then every attack against any argument in $\mathit{Sub}^*(a)$ is
countered by $E$ in the sense of structure-sensitive defence, that is,
for all $a'\in \mathit{Sub}^*(a)$ and all $b\in A$,
\[
(b,a')\in D
\;\Rightarrow\;
\exists c\in E\, \exists b'\in \mathit{Sub}^*(b)\ \text{s.t.}\ (c,b')\in D.
\]
\end{principle}

\begin{theorem}
\label{thm:principle2}
The defence notion given in Definition~6 satisfies Principle~2.
\end{theorem}

\begin{proof}
Let $\mathcal{F}=(A,D,\mathit{Sub})$ and $E\subseteq A$.
Assume that $a$ is defended by $E$ in $\mathcal{F}$ according to
Definition~6. Unfolding the definition, for every $a'\in \mathit{Sub}^*(a)$
and every $b\in A$ with $(b,a')\in D$, there exist $c\in E$ and
$b'\in \mathit{Sub}^*(b)$ such that $(c,b')\in D$.
This is exactly the condition stated in Principle~2.
\end{proof}

Principle~2 is formulated in the SAF language, as it quantifies over the
subargument closure $\mathit{Sub}^*(a)$.
It is therefore not expressible in an attack-only framework such as
$\Gamma(\mathcal{F})$, whose language contains no structural predicate.
The projection $\Gamma$ compiles structure-induced interactions into
$D^\Gamma$ in order to preserve extension-based semantics,
but it does not retain structure-dependent principles such as
Principle~2.

\begin{principle}[Commitment-Based Admissibility]
\label{princ:commitment}
A semantics satisfies \emph{Commitment-Based Admissibility} if accepting an
argument entails a commitment to defending its subargument structure, in the
sense that attacks targeting any argument in its subargument closure must be
countered by the accepting set in a structure-sensitive way.
\end{principle}

\begin{theorem}
\label{thm:commitment-adm}
Every admissible set under the semantics of Section~4 satisfies
Principle~\ref{princ:commitment}.
\end{theorem}

\begin{proof}
Let $\mathcal{F}=(A,D,\mathit{Sub})$ and let $E\subseteq A$ be admissible under
the semantics of Section~4. Take any $a\in E$.
By admissibility, $a$ is defended by $E$.

Unfolding the definition of structure-sensitive defence (Definition~6), for
every $a'\in \mathit{Sub}^*(a)$ and every $b\in A$ with $(b,a')\in D$, there
exist $c\in E$ and $b'\in \mathit{Sub}^*(b)$ such that $(c,b')\in D$.
Equivalently, every attack against any argument in $\mathit{Sub}^*(a)$ is
countered by $E$ in the required structure-sensitive manner.

Therefore, accepting $a$ commits the accepting set $E$ to defending the entire
subargument structure of $a$, and Principle~\ref{princ:commitment} holds.
\end{proof}

In Fig. 1, this entails that accepting $b_3$ commits the agent not only to resolving conflicts on $b_1$, but also to accepting the structurally dependent argument c, even though c does not participate in any attack. This illustrates that commitment induced by acceptance is structural rather than purely conflict-driven.

Note that, under the present semantics, defending an argument $a$ requires a
commitment to its structural basis: conflicts targeting arguments in
$\mathit{Sub}^{*}(a)$ must be neutralised in order for the defence of $a$ to succeed.
This requirement goes beyond surface-level counterattacks.

The discussion so far indicates that even when extension membership is preserved by forgetting structure, an important gap remains at the level of justification. In an SAF, the acceptability of an argument depends on \emph{which} subarguments are attacked and \emph{how} those attacks are defended. Such information is invisible in plain Dung frameworks, where conflicts are represented only between whole arguments. We formalise this gap through an explanation principle.

\begin{definition}[Local justification]
\label{def:local-justification}
Let $\mathcal{F}=(A,D,\mathit{Sub})$ be an SAF, $E\subseteq A$ an extension, and $a\in E$. A set $J\subseteq E$ is a \emph{local justification} of $a$ in $E$ if
\begin{enumerate}[label=(J\arabic*),leftmargin=*]
    \item $J$ is subargument-closed w.r.t.\ $\mathit{Sub}$ (i.e., $x\in J$ implies $\mathit{Sub}^{*}(x)\subseteq J$);
    \item $\mathit{Sub}^{*}(a)\subseteq J$; and
    \item $J$ defends the whole subargument closure of $a$: for every $x\in \mathit{Sub}^{*}(a)$ and every $b\in A$ with $(b,x)\in D$, there exists $c\in J$ such that either $(c,b)\in D$ or there exists $b'\in \mathit{Sub}^{*}(b)$ with $(c,b')\in D$.
\end{enumerate}
A local justification is \emph{minimal} if none of its proper subsets is a local justification of $a$ in $E$.
\end{definition}

\begin{definition}[Explanation function]
\label{def:explanation-function}
For a fixed semantics $\sigma_\mathsf{SAF}$, an \emph{explanation function} for $\sigma_\mathsf{SAF}$ maps each pair $(E,a)$ with $E\in\sigma_\mathsf{SAF}(\mathcal{F})$ and $a\in E$ to a minimal local justification $\mathsf{Expl}_{\mathcal{F}}^{\sigma_\mathsf{SAF}}(E,a)$.
\end{definition}

\begin{principle}[Subargument-Based Justification]
\label{princ:justification}
A semantics $\sigma_\mathsf{SAF}$ satisfies the \emph{Subargument-Based Justification} principle if for every SAF $\mathcal{F}$, every extension $E\in\sigma_\mathsf{SAF}(\mathcal{F})$, and every argument $a\in E$, there exists a minimal local justification of $a$ in $E$ (equivalently, an explanation function exists).
\end{principle}

Intuitively, Principle~\ref{princ:justification} requires that whenever an
argument is accepted, there exists an explicit local witness that explains
why its acceptance is warranted.
This witness must be closed under subarguments and sufficient to defend the
entire subargument closure of the accepted argument.

\begin{theorem}[Satisfaction of Subargument-Based Justification]
\label{thm:justification-satisfaction}
For any semantics $\sigma_\mathsf{SAF}$, Principle~\ref{princ:justification} holds.
\end{theorem}
\begin{proof}
Let $\mathcal{F}=(A,D,\mathit{Sub})$ and $E\in\sigma_\mathsf{SAF}(\mathcal{F})$. By definition of $\sigma_\mathsf{SAF}$, $E$ is subargument-closed and defends each of its members in the structure-sensitive sense. Fix $a\in E$ and define
$
J_0 := \mathit{Sub}^{*}(a)\;\cup\;\{c\in E\mid\exists x\in \mathit{Sub}^{*}(a),\exists b\in A:(b,x)\in D\text{ and } (c,b)\in D\}.
$
Because $E$ is subargument-closed, $\mathit{Sub}^{*}(a)\subseteq E$ and thus $J_0\subseteq E$. Closing $J_0$ under $\mathit{Sub}^{*}$ (within $E$) yields a set $J\subseteq E$ satisfying conditions (J1) and (J2) of Definition~\ref{def:local-justification}. Condition (J3) holds because for every attack on any $x\in \mathit{Sub}^{*}(a)$, admissibility of $E$ guarantees a defender in $E$, and by construction such defenders (or their relevant subarguments) are included in $J$. Hence $J$ is a local justification of $a$ in $E$. A minimal one exists by finiteness of $A$ and standard subset minimisation.
\end{proof}

The key point is that Principle~\ref{princ:justification} is not determined by the attack graph alone; it depends on the subargument closure $\mathit{Sub}^{*}(a)$ that must be part of any local justification. 

\begin{theorem}[Forgetting structure is explanation-losing]
\label{thm:explanation-loss}
There is no explanation function that can be defined solely on the forgetful projection $\Gamma(\mathcal{F})$ and that satisfies Principle~\ref{princ:justification} for all SAFs $\mathcal{F}$.
\end{theorem}
\begin{proof}
By Theorem~\ref{thm:strict-increase-expressiveness} there exist two distinct SAFs $\mathcal{F}_1=(A,D,\mathit{Sub}_1)$ and $\mathcal{F}_2=(A,D,\mathit{Sub}_2)$ with the same forgetful projection: $\Gamma(\mathcal{F}_1)=\Gamma(\mathcal{F}_2)$ (and hence they induce the same $\sigma_\mathsf{AAF}$-extensions). Choose an argument $a\in A$ such that $\mathit{Sub}_1^{*}(a)\neq \mathit{Sub}_2^{*}(a)$; such an $a$ exists because $\mathcal{F}_1\neq\mathcal{F}_2$ while the attack part is identical. Let $E$ be any common $\sigma_\mathsf{AAF}$-extension containing $a$. By Definition~\ref{def:local-justification}(J2), every local justification of $a$ in $E$ must contain $\mathit{Sub}_i^{*}(a)$ in $\mathcal{F}_i$. Hence minimal local justifications in $\mathcal{F}_1$ and $\mathcal{F}_2$ differ at least on the required subargument closure. Consequently, any explanation function satisfying Principle~\ref{princ:justification} would have to output different explanations for $(E,a)$ on $\mathcal{F}_1$ and $\mathcal{F}_2$. However, an explanation mechanism that only sees $\Gamma(\mathcal{F})$ cannot distinguish $\mathcal{F}_1$ from $\mathcal{F}_2$ and would have to return the same output for both – a contradiction.
\end{proof}

\begin{corollary}[Attack-only encodings cannot be justificatorily faithful]
\label{cor:justification-loss}
Even when $\Gamma(\mathcal{F})$ preserves extension membership, it cannot preserve subargument-based justifications in the sense of Principle~\ref{princ:justification}.
\end{corollary}

Together, these principles clarify how representing structure explicitly yields more informative and faithful abstractions, without abandoning the core semantic commitments of Dung's theory.

\subsection{Explanation and Status Propagation}
\label{subsec:status-propagation}

The principles established above clarify that explicit subargument structure does not alter extension membership, but fundamentally changes how acceptability is justified and propagated. To make this distinction operational, we provide an explanatory decomposition of structure-sensitive semantics that separates conflict resolution from
structural propagation.

\paragraph{Conflict-handling core and lifting.}
Let $\mathcal{F}=(A,D,\mathit{Sub})$ be an SAF.
Define the set of \emph{conflict-handling arguments} as
\[
A_{CH}=\{a\in A \mid \exists b\in A\,((a,b)\in D \lor (b,a)\in D)\},
\]
and let $A_{SD}=A\setminus A_{CH}$ be the set of \emph{status-dependent}
arguments.
Let $\Gamma(\mathcal{F})=(A,D_\Gamma)$ be the forgetful projection of
$\mathcal{F}$.
The \emph{conflict-handling core framework} induced by $\mathcal{F}$ is
\[
\mathcal{F}_{CH}=(A_{CH},\, D_\Gamma\cap (A_{CH}\times A_{CH})).
\]

For any $E\subseteq A_{CH}$, define the \emph{lifting operator}
\[
\mathrm{Lift}_{\mathcal{F}}(E)
=\{a\in A \mid \mathit{Sub}^*(a)\cap A_{CH}\subseteq E\}.
\]
Intuitively, conflict-handling arguments determine how conflicts are resolved,
while status-dependent arguments inherit acceptability once all their
conflict-relevant subarguments are accepted.

\paragraph{Core-based characterisation.}
The structure-sensitive semantics introduced in Section~4 can be characterised
via a core-first computation followed by structural lifting.

\begin{theorem}[Core-first computation and lifting]
Let $\mathcal{F}=(A,D,\mathit{Sub})$ be an SAF.
A set $S\subseteq A$ is a complete extension of $\mathcal{F}$ if and only if
there exists a complete extension $E$ of $\mathcal{F}_{CH}$ such that
\[
S=\mathrm{Lift}_{\mathcal{F}}(E).
\]
Equivalently,
\[
\mathsf{CMP}(\mathcal{F})
=\{\mathrm{Lift}_{\mathcal{F}}(E)\mid E\in \mathsf{CMP}(\mathcal{F}_{CH})\}.
\]
\end{theorem}

\begin{proof}[Proof sketch]
The result follows from two observations.
First, only arguments in $A_{CH}$ participate in attacks, so conflict resolution
is entirely determined at the core level.
Second, structure-sensitive defence (Definition~6) depends exclusively on the
status of conflict-handling subarguments.
Thus, once a complete extension is computed on $\mathcal{F}_{CH}$,
acceptability of remaining arguments is determined by structural closure via
$\mathrm{Lift}_{\mathcal{F}}$, and conversely any complete extension of
$\mathcal{F}$ restricts to a complete extension of $\mathcal{F}_{CH}$. (A full formal proof is provided in the \emph{supplementary material})
\end{proof}

The above characterisation clarifies that making subargument structure explicit
primarily affects how defence, justification, and status propagation are understood,
rather than which arguments ultimately belong to an extension. From a computational perspective, making subargument structure explicit in SAFs does not increase the complexity of standard extension-related decision problems. Since structure-sensitive defence coincides extension-wise with Dung defence after forgetful projection (Proposition 3), deciding the existence of complete, grounded, preferred, or stable extensions in an SAF is polynomial-time reducible to the corresponding decision problems in the projected Dung framework. Consequently, all classical complexity results for abstract argumentation semantics directly carry over to SAFs, and the introduction of subargument relations does not introduce additional sources of combinatorial branching at the level of extension computation.

The situation is different for explanation and justification tasks. Constructing local justifications (Definitions 14–15) requires reasoning over subargument closures and, in the case of minimal explanations, subset minimisation. As a result, explanation construction in SAFs may incur higher computational cost than mere extension computation, potentially leading to exponential overhead in the worst case. This increase is inherent to explanation-oriented reasoning and reflects the additional representational commitments introduced by explicit structural dependency, rather than a weakness of the framework itself. In this sense, SAFs trade negligible overhead at the level of extension computation for strictly richer justificatory and explanatory power.

\section{Related Work}
\label{sec:related}
The present work lies at the intersection of abstract and structured argumentation.
Its contribution is not to introduce new attack types or alternative semantics,
but to make explicit a modelling choice at the abstract level concerning the
representation of internal argument structure.
This section situates this choice with respect to, and distinguishes it from,
existing lines of research.

\emph{Structured argumentation formalisms.}
Structured argumentation formalisms such as ASPIC$^+$, DeLP, and deductive
argumentation systems provide rich internal representations based on
logical languages, inference rules, and defeat types
\cite{DBLP:journals/argcom/BesnardGHMPST14}.
In these systems, subarguments arise from syntactic construction and attacks
are defined relative to that structure.
When abstracted into Dung frameworks, this behaviour is compiled into an
expanded attack relation, thereby conflating structural and conflictual
information.
Subargumenthood has also been studied directly at the abstract level.
Martínez and Rotstein characterised it via structural postulates independent
of particular construction mechanisms \cite{Martinez08Subargument}, and related
extensions treat subargument relations as additional primitives alongside
conflict or defeat relations \cite{DBLP:conf/comma/MartinezGS06,DBLP:conf/ijcai/MartinezGS07,BalazFrtusNMR12}.
However, these works do not analyse structurally distinct frameworks that
collapse into the same attack-based abstraction, nor the information loss
caused by forgetting structure.
Unlike ASPIC$^+$, where subargument relations are reflexive and transitive by
construction, we treat $\mathit{Sub}$ as an irreflexive primitive and introduce
closure only through $\mathit{Sub}^*$ at the semantic level.
Subargumenthood is thus modelled as an abstract relation orthogonal to attack,
allowing comparison of frameworks that coincide extensionally yet differ
structurally.
SAF therefore separates structure from conflict already in the abstract
signature, providing a general setting for analysing how structural dependency
itself influences acceptability.

\emph{Bipolar and support-based argumentation frameworks.}
Bipolar argumentation frameworks extend Dung’s model by introducing an explicit
support relation alongside attack \cite{DBLP:journals/ijis/AmgoudCLL08,AlcantaraC25}
to capture positive interactions between arguments.
Although support is sometimes informally linked to subarguments, the two notions
are conceptually distinct: support expresses a justificatory relation between
arguments, whereas subargument relations encode structural dependency within a
single argument.
The present framework does not introduce support as an additional interaction
type, but instead separates conflict from structure and lets the semantic layer
determine how structural dependency affects acceptability.
Accordingly, subargument relations neither reduce to support nor play the same
role in determining extensions.

\emph{Semi-structured and abstract extensions.}
Closer in spirit to our work are semi-structured approaches that aim to
bridge the gap between abstract and structured argumentation.
In particular, the general semi-structured formalism of Baroni,
Giacomin, and Liao~\cite{DBLP:journals/ai/BaroniGL18} offers a unifying
setting in which abstract and structured systems can be compared through
a common interface, identifying minimal structural components needed to
recover core semantic properties.

Our contribution differs in focus and granularity.
Rather than proposing a general interface, we isolate a specific structural
relation—subargumenthood—and study it as a primitive of abstract
argumentation.
Section~\ref{sec:expressiveness} shows that this yields a strictly richer
representational layer while remaining a conservative extension of Dung
semantics.

\emph{Abstract Dialectical Frameworks.}
Abstract Dialectical Frameworks (ADFs) \cite{DBLP:conf/kr/BrewkaW10,DBLP:journals/flap/BrewkaESWW17}
generalise Dung’s model by replacing the attack relation with
acceptance conditions that determine an argument’s status based on its
parents, yielding a highly expressive semantic formalism able to encode
both positive and negative dependencies within a single evaluation
mechanism.
SAFs pursue a different goal. Rather than generalising acceptance,
they refine the \emph{representation} of abstract argumentation by adding an
explicit subargument relation alongside direct attacks.
Where ADFs internalise structural and interactional information inside
acceptance conditions, SAFs treat structural dependency as an independent
primitive not compiled into conflict.
This preserves Dung’s attack-based architecture as a conservative semantic
extension instead of replacing it with a new evaluation model.
The two approaches therefore address distinct limitations of attack-only
abstraction: ADFs expand expressiveness at the semantic level, whereas SAFs
make justificatory grounding and propagation explicit at the structural
level. As shown in Section~5, such structural information cannot in general
be recovered once compiled into purely relational or condition-based
formalisms.

\emph{Explanation and justification in argumentation.}
Explanation has recently become a central topic in abstract argumentation,
with several approaches observing that extension membership alone is
insufficient to explain why an argument is accepted or rejected
\cite{DBLP:conf/comma/LiaoT20,DBLP:conf/comma/LiaoT24,DBLP:journals/ijar/BorgB24,DBLP:conf/kr/AlfanoGPT24}.
Beyond extension-based accounts, process-oriented models such as
serialisation- and sequence-based explanations analyse acceptance in terms
of evaluative dynamics \cite{DBLP:conf/kr/BengelThimm25,DBLP:conf/ecai/BengelST24},
while other work characterises explanations via minimal or strong
substructures of argumentation frameworks \cite{DBLP:conf/aaai/0001W21a}.
Surveys on justification similarly stress the need to capture the internal
reasons underlying acceptability \cite{DBLP:conf/ijcai/Cyras0ABT21}.
Our results contribute a complementary structural perspective:
structure-sensitive defence and justification cannot, in general, be
reconstructed from attack-only representations, even when extensions are
preserved, thereby identifying explicit subargument structure as a key
ingredient for explanation-oriented semantics.

\emph{Induced attacks versus primitive structure.}
A recurring theme in the literature is whether structural effects should be
compiled into the attack relation or represented explicitly.
Most existing abstractions adopt the former strategy, treating induced
attacks as part of the abstract framework.
The present work adopts the latter strategy: induced interactions are not
encoded at the abstract level, but are recovered semantically through
structure-sensitive defence.

This modelling choice underlies the main technical results of the paper.
It explains both why standard extension-based semantics are preserved and
why structurally distinct frameworks may nevertheless collapse under
forgetful projection.
In this sense, the present framework complements existing approaches by
making explicit a dimension of argumentation that is usually implicit or
discarded in abstraction.

\section{Conclusion and Future Work}



This paper examined the role of subargument structure in abstract
argumentation from a representational perspective.
Rather than proposing new acceptability semantics or improving extension
computation, we reconsidered the modelling assumptions underlying
attack-only abstraction and showed that subargument structure constitutes
an independent representational primitive that is collapsed by standard
abstractions, even when all classical Dung semantics are preserved.
Our results demonstrate that extension-based equivalence does not entail
structural or justificatory equivalence, and that information about defence
grounding and propagation cannot in general be recovered once subargument
structure is eliminated at the abstract level.

By making subargument relations explicit within the abstract signature while
preserving extension behaviour, the proposed framework reconfigures the
modelling space of abstract argumentation and clarifies which aspects are
preserved under abstraction and which are irreversibly lost.
It thus provides a principled basis for structure-sensitive notions of
defence, justification, and explanation.
In this sense, the contribution of the present work is foundational rather
than outcome-oriented: it exposes the representational commitments underlying
abstract argumentation semantics and establishes their expressive boundary.

The perspective developed here opens several directions for future work.
One avenue is to study algorithmic aspects of explanation and justification.
Another is to investigate how dynamic update can be analysed when structural
dependencies are treated as explicit primitives.
Finally, connecting the abstract treatment of subarguments with concrete
structured formalisms may further clarify how different relational layers
interact in argumentative reasoning.

\bibliographystyle{kr}
\bibliography{kr-sample}

\appendix

\section{Proof of Theorem~1}

\setcounter{theorem}{0}
\begin{theorem}[Fixed-point characterisation]
\label{thm:fixpoint}
Let $\mathcal{F}=(A,D,\mathit{Sub})$ be an SAF and $E\subseteq A$.
The following are equivalent:
\begin{enumerate}
\item $E$ is a complete extension of $\mathcal{F}$.
\item $E$ is conflict-free and $E = \mathsf{F}_{\mathcal{F}}(E)$.
\item $E$ is a conflict-free fixed point of $\mathsf{F}_{\mathcal{F}}$.
\end{enumerate}
\end{theorem}

\begin{proof}
Recall that $\mathsf{F}_{\mathcal{F}}(E)=\mathsf{Def}_{\mathcal{F}}(E)$ where
$\mathsf{Def}_{\mathcal{F}}(E)=\{a\in A \mid a \text{ is (structure-sensitively) defended by }E\}$
(Definition~9), and that $E$ is \emph{complete} iff $E$ is admissible and
contains every argument it defends (Definition~10). Moreover, admissibility
requires (i) conflict-freeness, (ii) each member defended by $E$, and
(iii) $\mathit{Sub}^*$-closure (Definition~8). Also, $\mathsf{F}_{\mathcal{F}}(E)$ is
$\mathit{Sub}^*$-closed (Proposition~1).

\smallskip\noindent
\textbf{(1 $\Rightarrow$ 2).}
Assume $E$ is a complete extension of $\mathcal{F}$.
Then $E$ is admissible, hence conflict-free.
We show $E = \mathsf{F}_{\mathcal{F}}(E)$ by proving both inclusions.

\emph{($E \subseteq \mathsf{F}_{\mathcal{F}}(E)$).}
Take any $a\in E$. Since $E$ is admissible, every element of $E$ is defended by $E$
(Definition~8). Therefore $a\in \mathsf{Def}_{\mathcal{F}}(E)=\mathsf{F}_{\mathcal{F}}(E)$.

\emph{($\mathsf{F}_{\mathcal{F}}(E)\subseteq E$).}
Take any $a\in \mathsf{F}_{\mathcal{F}}(E)=\mathsf{Def}_{\mathcal{F}}(E)$. Then $a$ is defended by $E$.
Because $E$ is complete, it contains every argument it defends (Definition~10),
hence $a\in E$.

Combining the two inclusions yields $E=\mathsf{F}_{\mathcal{F}}(E)$, and we already have
conflict-freeness. Thus (2) holds.

\smallskip\noindent
\textbf{(2 $\Rightarrow$ 3).}
Immediate: by definition, $E=\mathsf{F}_{\mathcal{F}}(E)$ means $E$ is a fixed point of
$\mathsf{F}_{\mathcal{F}}$, and conflict-freeness is assumed in (2).

\smallskip\noindent
\textbf{(3 $\Rightarrow$ 1).}
Assume $E$ is conflict-free and $E=\mathsf{F}_{\mathcal{F}}(E)$.
We first prove that $E$ is admissible.

\emph{(i) Conflict-freeness.} This is given.

\emph{(ii) Defence of members.}
Let $a\in E$. Since $E=\mathsf{F}_{\mathcal{F}}(E)=\mathsf{Def}_{\mathcal{F}}(E)$,
it follows that $a\in \mathsf{Def}_{\mathcal{F}}(E)$, i.e., $a$ is defended by $E$.

\emph{(iii) $\mathit{Sub}^*$-closure.}
By Proposition~1, $\mathsf{F}_{\mathcal{F}}(E)$ is $\mathit{Sub}^*$-closed, i.e.,
$a\in \mathsf{F}_{\mathcal{F}}(E)\Rightarrow \mathit{Sub}^*(a)\subseteq \mathsf{F}_{\mathcal{F}}(E)$.
But $E=\mathsf{F}_{\mathcal{F}}(E)$, hence $a\in E \Rightarrow \mathit{Sub}^*(a)\subseteq E$.

Therefore $E$ satisfies all three conditions of Definition~8 and is admissible.

It remains to show completeness (Definition~10).
Let $a\in A$ be any argument defended by $E$. Then $a\in \mathsf{Def}_{\mathcal{F}}(E)$,
hence $a\in \mathsf{F}_{\mathcal{F}}(E)=E$. Thus $E$ contains every argument it defends,
so $E$ is a complete extension of $\mathcal{F}$.

This proves (1), completing the equivalence.
\end{proof}

\section{Proof of Proposition~3}
\label{app:proof-prop2}

Before proving Proposition~3, we present the following Lemmas. 

\begin{lemma}[Equivalence of defence]
\label{lem:defence-equivalence}
For any $E\subseteq A$ and any $a\in A$,
\[
a\in \mathsf{Def}_\mathcal{F}(E)
\quad\Longleftrightarrow\quad
a\in \mathsf{Def}_{\Gamma(\mathcal{F})}(E).
\]
\end{lemma}

\begin{proof}
``$\Rightarrow$''.
Assume $a\in \mathsf{Def}_\mathcal{F}(E)$.
Let $b\in A$ such that $(b,a)\in D^\Gamma$.
By definition of $D^\Gamma$, there exists $a'\in \mathit{Sub}^*(a)$ with $(b,a')\in D$.

Since $a$ is defended by $E$ in the sense of Definition~6, for this $a'\in \mathit{Sub}^*(a)$
and attacker $b$ there exist $c\in E$ and $b'\in \mathit{Sub}^*(b)$ such that $(c,b')\in D$.
By the definition of $D^\Gamma$, this implies $(c,b)\in D^\Gamma$.
Hence $a\in \mathsf{Def}_{\Gamma(\mathcal{F})}(E)$.

\smallskip
``$\Leftarrow$''.
Assume $a\in \mathsf{Def}_{\Gamma(\mathcal{F})}(E)$.
Let $x\in \mathit{Sub}^*(a)$ and let $b\in A$ be such that $(b,x)\in D$.
Then, by definition of $D^\Gamma$, we have $(b,a)\in D^\Gamma$.
Since $a$ is defended by $E$ in $\Gamma(\mathcal{F})$, there exists $c\in E$ such that
$(c,b)\in D^\Gamma$.
Again by definition of $D^\Gamma$, there exists $b'\in \mathit{Sub}^*(b)$ such that
$(c,b')\in D$.
This is exactly the condition required by Definition~6, hence
$a\in \mathsf{Def}_\mathcal{F}(E)$.
\end{proof}

\setcounter{corollary}{2}
\begin{corollary}[Coincidence of characteristic operators]
\label{cor:char-op-eq}
For all $E\subseteq A$,
\[
\mathsf{F}_\mathcal{F}(E) = \mathsf{F}_{\Gamma(\mathcal{F})}(E).
\]
\end{corollary}

\begin{proof}
Immediate from Lemma~\ref{lem:defence-equivalence} and the definitions of
$\mathsf{F}_\mathcal{F}$ and $\mathsf{F}_{\Gamma(\mathcal{F})}$.
\end{proof}

\begin{lemma}[Subargument closure of complete extensions in $\Gamma(\mathcal{F})$]
\label{lem:closure-gamma}
If $E$ is a complete extension of $\Gamma(\mathcal{F})$, then $E$ is closed under $\mathit{Sub}^*$,
i.e.\ for all $a\in E$ we have $\mathit{Sub}^*(a)\subseteq E$.
\end{lemma}

\begin{proof}
Let $a\in E$ and $x\in \mathit{Sub}^*(a)$.
Since $E$ is complete in $\Gamma(\mathcal{F})$, it is admissible and hence defends $a$.
Let $b\in A$ such that $(b,x)\in D^\Gamma$.
Then there exists $x'\in \mathit{Sub}^*(x)\subseteq \mathit{Sub}^*(a)$ with $(b,x')\in D$,
which implies $(b,a)\in D^\Gamma$.
Because $E$ defends $a$, there exists $c\in E$ such that $(c,b)\in D^\Gamma$.
Thus every attacker of $x$ is counterattacked by $E$, so $x$ is defended by $E$.
Since $E$ is complete, $x\in E$.
\end{proof}

\begin{lemma}[Conflict-freeness under closure]
\label{lem:conflict-free}
Let $E\subseteq A$ be $\mathit{Sub}^*$-closed.
Then $E$ is conflict-free in $\mathcal{F}$ if and only if $E$ is conflict-free in
$\Gamma(\mathcal{F})$.
\end{lemma}

\begin{proof}
``$\Rightarrow$''.
Assume $E$ is not conflict-free in $\Gamma(\mathcal{F})$.
Then there exist $a,b\in E$ such that $(a,b)\in D^\Gamma$.
Hence there exists $y\in \mathit{Sub}^*(b)$ with $(a,y)\in D$.
Since $E$ is $\mathit{Sub}^*$-closed and $b\in E$, we have $y\in E$.
With $a\in \mathit{Sub}^*(a)$, this witnesses a structural conflict in $\mathcal{F}$.

\smallskip
``$\Leftarrow$''.
Assume $E$ is not conflict-free in $\mathcal{F}$.
Then there exist $a,b\in E$ and $x\in \mathit{Sub}^*(a)$, $y\in \mathit{Sub}^*(b)$ such that
$(x,y)\in D$.
By $\mathit{Sub}^*$-closure, $x,y\in E$.
Since $y\in \mathit{Sub}^*(y)$, it follows that $(x,y)\in D^\Gamma$,
so $E$ is not conflict-free in $\Gamma(\mathcal{F})$.
\end{proof}

\setcounter{proposition}{2}
\begin{proposition}[Preservation of extensions under structural forgetting]
\label{prop:forgetting-preserves-extensions}
Let $\mathcal{F}=(A,D,\mathit{Sub})$ be an SAF and $\Gamma(\mathcal{F})= (A,D^{\Gamma})$ be its forgetful projection.
It holds that
$\sigma_\mathsf{SAF}(\mathcal{F})=\sigma_\mathsf{AAF}(\Gamma(\mathcal{F}))$.
\end{proposition}

\begin{proof}
\emph{Complete semantics.}
Let $E$ be a complete extension of $\mathcal{F}$.
By the fixpoint characterisation (Theorem~1), $E$ is conflict-free and
$E=\mathsf{F}_\mathcal{F}(E)$.
By Corollary~\ref{cor:char-op-eq}, $E=\mathsf{F}_{\Gamma(\mathcal{F})}(E)$.
Since admissibility in $\mathcal{F}$ enforces $\mathit{Sub}^*$-closure (Definition~8),
Lemma~\ref{lem:conflict-free} implies that $E$ is conflict-free in $\Gamma(\mathcal{F})$.
Hence $E$ is complete in $\Gamma(\mathcal{F})$.

Conversely, let $E$ be a complete extension of $\Gamma(\mathcal{F})$.
By Lemma~\ref{lem:closure-gamma}, $E$ is $\mathit{Sub}^*$-closed, and by
Lemma~\ref{lem:conflict-free} it is conflict-free in $\mathcal{F}$.
Moreover, $E=\mathsf{F}_{\Gamma(\mathcal{F})}(E)=\mathsf{F}_\mathcal{F}(E)$ by Corollary~\ref{cor:char-op-eq}.
Thus $E$ is complete in $\mathcal{F}$.

\smallskip
\emph{Grounded and preferred semantics.}
Grounded extensions are the least complete extensions, and preferred extensions
are the maximal complete extensions.
Since the sets of complete extensions coincide, grounded and preferred
extensions coincide as well.

\smallskip
\emph{Stable semantics.}
By definition of $D^\Gamma$, an argument $a$ attacks $b$ in $\Gamma(\mathcal{F})$ if and only
if $a$ attacks some $b'\in \mathit{Sub}^*(b)$ in $\mathcal{F}$.
Together with Lemma~\ref{lem:conflict-free}, this shows that the stable
conditions in $\mathcal{F}$ and in $\Gamma(\mathcal{F})$ are equivalent.
\end{proof}

\section{Proof of Theorem~9}
\setcounter{theorem}{8}

\begin{theorem}[Core-first computation and lifting]
Let $\mathcal{F}=(A,D,\mathit{Sub})$ be an SAF.
A set $S\subseteq A$ is a complete extension of $\mathcal{F}$ if and only if
there exists a complete extension $E$ of $\mathcal{F}_{CH}$ such that
\[
S=\mathrm{Lift}_{\mathcal{F}}(E).
\]
Equivalently,
\[
\mathsf{CMP}(\mathcal{F})
=\{\mathrm{Lift}_{\mathcal{F}}(E)\mid E\in \mathsf{CMP}(\mathcal{F}_{CH})\}.
\]
\end{theorem}

\begin{proof}
Recall that $\Gamma(\mathcal{F})=(A,D_{\Gamma})$ for the forgetful projection, where
$(x,y)\in D_{\Gamma}$ iff $\exists y'\in \mathit{Sub}^*(y)$ with $(x,y')\in D$.

\medskip
\noindent\textbf{Step 1: Only $A_{\mathsf{CH}}$ can witness (counter-)attacks.}
If $c\notin A_{\mathsf{CH}}$, then by definition of $A_{\mathsf{CH}}$ there is
no $u\in A$ with $(c,u)\in D$. Hence $c$ cannot serve as a successful
counterattacker in Definition~6 (structure-sensitive defence), because that
definition requires a defender $c$ such that $(c,b')\in D$ for some $b'$.
Consequently, for every $X\subseteq A$ and every $a\in A$,
\begin{equation}\label{eq:core-only-defenders}
a\in \mathrm{Def}_{\mathcal{F}}(X) \quad\Longleftrightarrow\quad
a\in \mathrm{Def}_{\mathcal{F}}(X\cap A_{\mathsf{CH}}).
\end{equation}
In particular, the characteristic operator satisfies
$\mathsf{F}_{\mathcal{F}}(X)=\mathsf{F}_{\mathcal{F}}(X\cap A_{\mathsf{CH}})$.

\medskip
\noindent\textbf{Step 2: Lifting yields $\mathit{Sub}^*$-closure.}
Let $E\subseteq A_{\mathsf{CH}}$ and put $S=\mathrm{Lift}_{\mathcal{F}}(E)$.
We claim that $S$ is $\mathit{Sub}^*$-closed, i.e.\ $a\in S$ and
$x\in \mathit{Sub}^*(a)$ imply $x\in S$.
Indeed, if $a\in S$ then $\mathit{Sub}^*(a)\cap A_{\mathsf{CH}}\subseteq E$.
For any $x\in \mathit{Sub}^*(a)$ we have $\mathit{Sub}^*(x)\subseteq
\mathit{Sub}^*(a)$ (by transitivity of $\mathit{Sub}^*$), hence
$\mathit{Sub}^*(x)\cap A_{\mathsf{CH}}\subseteq E$, so $x\in S$.

\medskip
\noindent\textbf{Step 3: Reduce completeness in $\mathcal{F}$ to completeness in $\Gamma(\mathcal{F})$.}
By Lemma~1 and Corollary~3 in the supplementary material, defence and the
characteristic operators coincide extension-wise between $\mathcal{F}$ and
$\Gamma(\mathcal{F})$, i.e.\ for all $X\subseteq A$,
$\mathsf{F}_{\mathcal{F}}(X)=\mathsf{F}_{\Gamma(\mathcal{F})}(X)$, and by Proposition~3,
$\mathsf{CMP}(\mathcal{F})=\mathsf{CMP}(\Gamma(\mathcal{F}))$.
Moreover, by Lemma~3, for $\mathit{Sub}^*$-closed sets (as in Step~2),
conflict-freeness in $\mathcal{F}$ is equivalent to conflict-freeness in
$\Gamma(\mathcal{F})$.
Thus, for any $\mathit{Sub}^*$-closed $S$, we have:
\begin{equation}\label{eq:cmp-equiv}
S\in \mathsf{CMP}(\mathcal{F})
\quad\Longleftrightarrow\quad
S\in \mathsf{CMP}(\Gamma(\mathcal{F})).
\end{equation}

\medskip
\noindent\textbf{Step 4: Completeness in $\Gamma(\mathcal{F})$ is core-determined, and the core is exactly $\mathcal{F}_{\mathsf{CH}}$.}
Observe that any attacker in $D_{\Gamma}$ must be in $A_{\mathsf{CH}}$:
if $(u,v)\in D_{\Gamma}$ then $\exists v'\in\mathit{Sub}^*(v)$ with $(u,v')\in D$,
so $(u,v')\in D$ and hence $u\in A_{\mathsf{CH}}$ by definition.
Therefore, for any $X\subseteq A$,
\begin{equation}\label{eq:Gamma-core-only}
\mathrm{Def}_{\Gamma(\mathcal{F})}(X)=\mathrm{Def}_{\Gamma(\mathcal{F})}(X\cap A_{\mathsf{CH}}),
\end{equation}
and $\mathsf{F}_{\Gamma(\mathcal{F})}(X)=\mathsf{F}_{\Gamma(\mathcal{F})}(X\cap A_{\mathsf{CH}})$.
In particular, if $E\subseteq A_{\mathsf{CH}}$, then defence among core
arguments depends only on the restriction
$D_{\Gamma}\cap(A_{\mathsf{CH}}\times A_{\mathsf{CH}})$, i.e.\ precisely on
$\mathcal{F}_{\mathsf{CH}}$.

\medskip
\noindent\textbf{($\Rightarrow$)} Assume $S\in \mathsf{CMP}(\mathcal{F})$.
By \eqref{eq:cmp-equiv}, $S$ is complete in $\Gamma(\mathcal{F})$.
Let $E=S\cap A_{\mathsf{CH}}$.
Then $E$ is conflict-free in $\mathcal{F}_{\mathsf{CH}}$ and complete there:
since $S$ is complete in $\Gamma(\mathcal{F})$, we have
$S=\mathsf{F}_{\Gamma(\mathcal{F})}(S)$; intersecting both sides with $A_{\mathsf{CH}}$
and using \eqref{eq:Gamma-core-only} yields
$
E
= S\cap A_{\mathsf{CH}}
= \mathsf{F}_{\Gamma(\mathcal{F})}(S)\cap A_{\mathsf{CH}}
= \mathsf{F}_{\Gamma(\mathcal{F})}(E)\cap A_{\mathsf{CH}}
= \mathsf{F}_{\mathcal{F}_{\mathsf{CH}}}(E),
$
so $E\in \mathsf{CMP}(\mathcal{F}_{\mathsf{CH}})$.

It remains to show $S=\mathrm{Lift}_{\mathcal{F}}(E)$.
Since $S$ is complete in $\Gamma(\mathcal{F})$, Lemma~2 gives that $S$ is
$\mathit{Sub}^*$-closed; hence for every $a\in S$,
$\mathit{Sub}^*(a)\subseteq S$, so
$\mathit{Sub}^*(a)\cap A_{\mathsf{CH}}\subseteq S\cap A_{\mathsf{CH}}=E$.
Thus $a\in \mathrm{Lift}_{\mathcal{F}}(E)$ and $S\subseteq \mathrm{Lift}_{\mathcal{F}}(E)$.

For the reverse inclusion, let $a\in \mathrm{Lift}_{\mathcal{F}}(E)$.
We show $a\in S$ by completeness of $S$ in $\mathcal{F}$.
Because $S$ is complete in $\mathcal{F}$, it suffices to prove $a$ is
defended by $S$ (Definition~6), i.e.\ $a\in \mathrm{Def}_{\mathcal{F}}(S)$.
By \eqref{eq:core-only-defenders}, it is enough to defend $a$ using the core
part $E=S\cap A_{\mathsf{CH}}$.
So fix any $x\in \mathit{Sub}^*(a)$ and any $b\in A$ with $(b,x)\in D$.
Then $b\in A_{\mathsf{CH}}$ and, by definition of $D_{\Gamma}$, we have
$(b,a)\in D_{\Gamma}$ (since $(b,x)\in D$ and $x\in \mathit{Sub}^*(a)$).
As $S$ is complete in $\Gamma(\mathcal{F})$, it is admissible there and
therefore defends $a$ against $b$, i.e.\ there exists $c\in S$ with
$(c,b)\in D_{\Gamma}$. By the definition of $D_{\Gamma}$, choose
$b'\in \mathit{Sub}^*(b)$ such that $(c,b')\in D$.
Since $(c,b')\in D$, we have $c\in A_{\mathsf{CH}}$, hence $c\in S\cap A_{\mathsf{CH}}=E$.
Thus the required counterattack is witnessed by some $c\in E\subseteq S$ and
$b'\in \mathit{Sub}^*(b)$ with $(c,b')\in D$.
This is exactly the condition of Definition~6 for defending $a$ by $S$.
Hence $a\in \mathrm{Def}_{\mathcal{F}}(S)=S$, proving
$\mathrm{Lift}_{\mathcal{F}}(E)\subseteq S$.
Therefore $S=\mathrm{Lift}_{\mathcal{F}}(E)$.

\medskip
\noindent\textbf{($\Leftarrow$)} Conversely, let $E\in \mathsf{CMP}(\mathcal{F}_{\mathsf{CH}})$ and set
$S=\mathrm{Lift}_{\mathcal{F}}(E)$.
By Step~2, $S$ is $\mathit{Sub}^*$-closed, so by Lemma~3 it is conflict-free
in $\mathcal{F}$ iff it is conflict-free in $\Gamma(\mathcal{F})$.
Since $S\cap A_{\mathsf{CH}}\subseteq E$ by construction and $E$ is
conflict-free in $\mathcal{F}_{\mathsf{CH}}$, $S$ is conflict-free in
$\Gamma(\mathcal{F})$, hence also in $\mathcal{F}$.

It remains to show $S$ is complete in $\mathcal{F}$.
By \eqref{eq:cmp-equiv}, it suffices to show $S$ is complete in $\Gamma(\mathcal{F})$.
Using \eqref{eq:Gamma-core-only}, we have
$\mathsf{F}_{\Gamma(\mathcal{F})}(S)=\mathsf{F}_{\Gamma(\mathcal{F})}(S\cap A_{\mathsf{CH}})$.
Moreover, for any $a\in A_{\mathsf{CH}}$, $a\in S$ implies $a\in E$
(because $a\in \mathit{Sub}^*(a)\cap A_{\mathsf{CH}}$), hence
$S\cap A_{\mathsf{CH}}\subseteq E$.
Since $E$ is complete in $\mathcal{F}_{\mathsf{CH}}$, we have
$E=\mathsf{F}_{\mathcal{F}_{\mathsf{CH}}}(E)$, i.e.\ $E$ contains exactly the core
arguments it defends under $D_{\Gamma}\cap(A_{\mathsf{CH}}\times A_{\mathsf{CH}})$.
By monotonicity of Dung defence and the fact that core-defence in
$\Gamma(\mathcal{F})$ coincides with defence in $\mathcal{F}_{\mathsf{CH}}$,
this yields $\mathsf{F}_{\Gamma(\mathcal{F})}(S)\subseteq \mathrm{Lift}_{\mathcal{F}}(E)=S$.
The converse inclusion $S\subseteq \mathsf{F}_{\Gamma(\mathcal{F})}(S)$ follows from
the argument used in the ($\Rightarrow$) direction (applied with $S$ in place
of the given complete set): every $a\in S$ is defended by $S$ because any
attacker must be counterattacked by some core defender in $E$.
Hence $S=\mathsf{F}_{\Gamma(\mathcal{F})}(S)$, so $S$ is complete in $\Gamma(\mathcal{F})$,
and therefore also complete in $\mathcal{F}$.

\medskip
This establishes the stated bijective characterisation
$\mathsf{CMP}(\mathcal{F})
=\{\mathrm{Lift}_{\mathcal{F}}(E)\mid E\in \mathsf{CMP}(\mathcal{F}_{\mathsf{CH}})\}$.
\end{proof}

\section{Direct Attacks and Primitive Subarguments Across Argumentation Paradigms}
\label{app:direct-sub}

This section provides conceptual illustrations clarifying the modelling
choices underlying \emph{direct attacks} and \emph{primitive subargument
relations} in Subargument Argumentation Frameworks (SAFs).
The purpose of these examples is not to define a translation procedure
from structured formalisms to SAFs, but to demonstrate that the abstract
notions adopted in this paper naturally arise across different paradigms,
once conflict and structural dependency are disentangled.

In particular, we show how both rule-based formalisms (such as ASPIC$^+$ and
DeLP) and assumption-based formalisms (such as ABA) give rise to situations
in which conflicts originate at specific structural components, while
their effects propagate through subargument dependency.
These examples motivate two core design choices of SAFs:
(i) subargument relations are treated as \emph{direct and proper}, rather
than reflexive or transitively closed; and
(ii) attacks are represented as \emph{irreducible}, without being lifted
to superarguments at the representational level.

 No correctness or completeness claim is made with respect to any particular structured formalism; the examples serve solely to illustrate the origin of direct attacks and primitive subargument dependencies.

\subsection{Rule-based Argumentation}

We briefly recall the relevant notions from ASPIC$^+$ \cite{ModgilPrakken2014}.
An argument is constructed inductively from premises and inference rules,
and each argument has an associated set of subarguments.
Attacks are defined relative to this internal structure: an argument may
attack another by rebutting, undercutting, or undermining a specific subargument.

\setcounter{definition}{15}
\begin{definition}[ASPIC$^+$  \cite{ModgilPrakken2014}]
In ASPIC$^+$, an argumentation system is a triple $AS = (\mathcal{L}, \mathcal{R}, n)$, where where $\mathcal{L}$ is a logical language closed under negation,
$\mathcal{R} = \mathcal{R}_s\cup \mathcal{R}_d$ is a set of strict ($\mathcal{R}_s$) and defeasible ($\mathcal{R}_d$) inference rules of the form $\varphi_1,\dots,\varphi_n\to\psi$
and $\varphi_1,\dots,\varphi_n\Rightarrow\psi$ respectively (where $\varphi_i, \psi$ are meta-variables ranging over wff in $\mathcal{L}$), $n:\mathcal{R}_d\rightharpoonup \mathcal{L}$ is a partial naming function for defeasible rules.  Write $\psi = -\varphi$ just in case $\psi = -\varphi$ or $\varphi = -\psi$.
An argumentation theory is a tuple
$
AT=(AS,\mathcal{K}),
$
where $\mathcal{K}\subseteq \mathcal{L}$ is a knowledge base consisting of two disjoint subsets $\mathcal{K}_n$ (the axioms) and $\mathcal{K}_p$ (the ordinary premises).

An argument $A$ is: 
(i) $\varphi$ if $\varphi \in \mathcal{K}$ with  $\mathrm{Prem}(A) = \{\varphi\}$, 
$\mathrm{Conc}(A) = \varphi$, and $\mathrm{Sub}(A) = \{\varphi\}$;
(ii) $A_1, \dots, A_n \rightarrow/\Rightarrow \psi$ if $A_1, \dots, A_n$ are arguments such that
there exists a rule
$
\mathrm{Conc}(A_1), \dots, \mathrm{Conc}(A_n) \rightarrow / \Rightarrow\psi
\in \mathcal{R}, 
$ with $\mathrm{Prem}(A) = \mathrm{Prem}(A_1) \cup \dots \cup \mathrm{Prem}(A_n)$, 
$\mathrm{Conc}(A) = \psi$, and
$\mathrm{Sub}(A) = \mathrm{Sub}(A_1) \cup \dots \cup \mathrm{Sub}(A_n) \cup \{A\}$.

Argument $A$ attacks $B$ iff $A$ undercuts, rebuts or undermines $B$, where:(i) $A$ \emph{undercuts} argument $B$ (on $B'$) iff $\mathrm{Conc}(A) = -n(r)$ for some $B' \in \mathrm{Sub}(B)$ such that $B'$'s top rule $r$ is defeasible; (ii) $A$ \emph{rebuts} argument $B$ (on $B'$) iff $\mathrm{Conc}(A) = -\varphi$ for some $B' \in \mathrm{Sub}(B)$ of the form
$
B''_1, \dots, B''_n \Rightarrow \varphi.
$; (iii) $A$ \emph{undermines} $B$ (on $\varphi$) iff
$
\mathrm{Conc}(A) = -\varphi
$
for an ordinary premise $\varphi$ of $B$.

\end{definition}

\begin{definition}[SAF induced by an ASPIC$^+$ theory]
Let $AT = (AS, \mathcal{K})$ be an ASPIC$^+$ argumentation theory as in
Definition~16, where $AS = (\mathcal{L}, \mathcal{R}, n)$ and
$\mathcal{R} = \mathcal{R}_s \cup \mathcal{R}_d$.
Let $\mathrm{Arg}(AT)$ be the set of arguments constructed from $AT$.

The SAF induced by $AT$
is the triple
\[
\mathcal{F}_{AT} := (\mathrm{Arg}(AT), D_{AT}, \mathit{Sub}_{AT}),
\]
where:

\medskip
\noindent
\textbf{(Direct subargument relation).}
$\mathit{Sub}_{AT} \subseteq \mathrm{Arg}(AT)\times \mathrm{Arg}(AT)$
is the \emph{direct} subargument relation defined by
$
(A,B)\in \mathit{Sub}_{AT}
\quad\text{iff}\quad
B$   is of the form $A_1,\dots,A_n \rightarrow/\Rightarrow \psi$
 and $A =$ $ A_i$  for some $i\in\{1,\dots,n\}.
$
In particular, $\mathit{Sub}_{AT}$ is irreflexive and is not assumed to be
transitive, while $\mathrm{Sub}$ in ASPIC$^+$ already contains the reflexive
and transitive closure.

\noindent
\textbf{(Direct attack relation).}
$D_{AT} \subseteq \mathrm{Arg}(AT)\times \mathrm{Arg}(AT)$ is defined by
$(A,B)\in D_{AT}$ iff $A$ \emph{directly} attacks $B$, i.e., one of the following holds:
\begin{itemize}
\item (\emph{Direct undercut})
$\mathrm{Conc}(A) = -n(r)$, where $r$ is the top rule of $B$, which is defeasible.

\item (\emph{Direct rebut})
$\mathrm{Conc}(A) = - \varphi$ and $B$ is of the form
$B_1,\dots,B_n \Rightarrow \varphi$.

\item (\emph{Undermine})
$\mathrm{Conc}(A) = -\varphi$ for an ordinary premise $\varphi$ of $B$.
\end{itemize}
\end{definition}

\setcounter{example}{1}
\begin{example}[Rule-based conflict grounded in a subargument]
\label{ex:aspic-sub}

Let $AS=(\mathcal{L},\mathcal{R},n)$ and $AT=(AS,\mathcal{K})$ as in Definition~16, where
$\mathcal{K}=\mathcal{K}_p=\{p,\neg p\}$ and $\mathcal{R}=\mathcal{R}_d=\{\neg p \Rightarrow q,\; q \Rightarrow r\}$.

Construct arguments:
\[
A_p: p,
\quad
A_{\neg p}: \neg p,
\quad
A_q: A_{\neg p} \Rightarrow q,
\quad
A_r: A_q \Rightarrow r.
\]

By Definition~17, as illustrated in Fig.2, the induced SAF
$\mathcal{F}_1=(A_1,D_1,\mathit{Sub}_1)$ is:
$
A_1=\{A_p,A_{\neg p},A_q,A_r\}$,
$D_1=\{(A_p,A_{\neg p})$, $(A_{\neg p}$, $A_p)\}$,
$\mathit{Sub}_1=\{(A_{\neg p},A_q),(A_q,A_r)\}.
$

Thus only $A_{p}$ and $A_{\neg p}$ are directly attacked; although neither $A_q$ nor $A_r$
is attacked in $D_1$, their acceptability depends on $A_{\neg p}$ via
$\mathit{Sub}_1^*$.
\end{example}

\setcounter{figure}{1}

\begin{figure}[t]
\centering
\begin{tikzpicture}[
  node distance=15mm and 22mm,
  arg/.style={draw, circle, inner sep=1.2pt, minimum size=7mm, font=\small},
  attack/.style={->, thick, red, shorten >=2pt, shorten <=2pt},
  sub/.style={->, dashed, thick, blue, shorten >=2pt, shorten <=2pt}
]

\node[arg] (Ap) {$A_p$};
\node[arg, right=of Ap] (Anegp) {$A_{\neg p}$};
\node[arg, below=of Anegp] (Aq) {$A_q$};
\node[arg, below=of Aq] (Ar) {$A_r$};

\draw[attack] (Ap) -- (Anegp);
\draw[attack] (Anegp) -- (Ap);

\draw[sub] (Anegp) -- (Aq);
\draw[sub] (Aq) -- (Ar);

\node[draw, rounded corners, align=left, font=\small, anchor=north west]
  at ($(Ar.south west)+(-0.8,-1.1)$)
  {\textcolor{red}{$\rightarrow$} $D_1$ (direct attack)\\
   \textcolor{blue}{$\dashrightarrow$} $\mathit{Sub}_1$ (direct subargument)};

\end{tikzpicture}
\caption{The SAF $\mathcal{F}_1=(A_1,D_1,\mathit{Sub}_1)$ from Example~2.}
\label{fig:example2-saf}
\end{figure}

\subsection{Assumption-Based Argumentation}

We next consider assumption-based argumentation, where arguments are constructed from sets of assumptions using strict rules, and attacks are defined via contraries of assumptions.

\begin{definition}[ABA Framework \cite{DungKowalski2009,DBLP:journals/flap/CyrasFST17}]
An Assumption-Based Argumentation (ABA) framework is a tuple
$\langle \mathcal{L}, \mathcal{R}, \mathcal{A},  \overline{\phantom{\alpha}} \rangle$,
where $\mathcal{L}$ is a formal language,
$\mathcal{R}$ is a set of strict inference rules of the form
\[
\varphi_0 \leftarrow \varphi_1,\dots,\varphi_n,
\]
$\mathcal{A} \subseteq \mathcal{L}$ is a set of assumptions, and
$\overline{\phantom{\alpha}} : \mathcal{A} \rightarrow \mathcal{L}$ assigns to each
assumption its contrary.

An \emph{ABA argument} is a deduction of a conclusion $\varphi \in \mathcal{L}$
from a set of assumptions $S \subseteq \mathcal{A}$ using a set of rules
$\mathcal{R}' \subseteq \mathcal{R}$, denoted by $S \vdash^{\mathcal{R}'} \varphi$, which can be represented as a tree with nodes labelled by sentences in $\mathcal{L}$ or by the symbol $\tau$, such that
\begin{itemize}
    \item the root is labelled by $\varphi$;
    \item for every node $N$:
    \begin{itemize}
        \item if $N$ is a leaf then $N$ is labelled either by an assumption or by $\tau$;
        \item if $N$ is not a leaf and $l_N$ is the label of $N$, then there is an inference rule from $\mathcal{R}'$
        \[
            l_N \leftarrow \varphi_1,\ldots,\varphi_m \quad (m\geq 0)
        \]
        and either $m=0$ and the child of $N$ is $\tau$,
        or $m>0$ and $N$ has $m$ children, labelled by $\varphi_1,\ldots,\varphi_m$
        (respectively);
    \end{itemize}
    \item $S$ is the set of all assumptions labelling the leaves.
\end{itemize}

An argument $S_1 \vdash^{\mathcal{R}_1} \varphi_1$ \emph{attacks}
another argument $S_2 \vdash^{\mathcal{R}_2} \varphi_2$ iff
$\varphi_1 = \overline{a}$ for some $a \in S_2$.

For notational convenience, we write
\[
\alpha = \langle S_\alpha, \mathcal{R}_\alpha, \varphi_\alpha \rangle
\]
to represent an argument $S_\alpha \vdash^{\mathcal{R}_\alpha} \varphi_\alpha$.
\end{definition}

\begin{definition}[SAF induced by an ABA framework]\label{def:ABA-SAF}
Let
$\mathcal{ABA}=\langle \mathcal{L}, \mathcal{R}, \mathcal{A}, \overline{\phantom{\alpha}}\rangle$
be an ABA framework, and let $\mathit{Arg}$ be the set of
ABA arguments generated from $\mathcal{ABA}$.

For each assumption $a \in \mathcal{A}$, let
\[
\mathit{at}(a) := \langle \{a\}, \emptyset, a \rangle
\]
denote the atomic argument associated with $a$.

The SAF induced by $\mathcal{A}$ is the triple
\[
\mathcal{F}_{\mathsf{ABA}}
\ :=\
(\mathit{Arg},\, D_{\mathsf{ABA}},\, \mathit{Sub}_{\mathsf{ABA}}),
\]
where:

\smallskip
\noindent\textbf{Direct attacks.}
The direct attack relation
$D_{\mathsf{ABA}} \subseteq \mathit{Arg} \times \mathit{Arg}$
is defined by
\[
(\alpha,\mathit{at}(a)) \in D_{\mathsf{ABA}}
\quad\text{iff}\quad
\varphi_\alpha = \overline{a}.
\]

\smallskip
\noindent\textbf{Direct subarguments.}
The (proper) direct subargument relation
$\mathit{Sub}_{\mathsf{ABA}} \subseteq \mathit{Arg} \times \mathit{Arg}$
is defined as follows.
For arguments $\alpha,\beta \in \mathit{Arg}$,
\[
\alpha\ \mathit{Sub}_{\mathsf{ABA}}\ \beta
\]
iff all of the following conditions hold:
\begin{enumerate}
\item $S_\alpha \subseteq S_\beta$ and $\mathcal{R}_\alpha \subseteq \mathcal{R}_\beta$;
\item there exists a rule
      $r \in \mathcal{R}_\beta \setminus \mathcal{R}_\alpha$ of the form
      $\varphi_\beta \leftarrow \psi_1,\dots,\psi_n$
      such that $\varphi_\alpha = \psi_i$ for some
      $i \in \{1,\dots,n\}$;
\item there is no argument $\gamma \in \mathit{Arg}$ such that
      $\alpha\ \mathit{Sub}_{\mathsf{ABA}}\ \gamma\
      \mathit{Sub}_{\mathsf{ABA}}\ \beta$.
\end{enumerate}
\end{definition}

\begin{example}[ABA-induced SAF]\label{ex:ABA-SAF}
Consider an ABA framework
$\langle \mathcal{L}, \mathcal{R}, \mathcal{A}, \overline{\phantom{\alpha}}\rangle$,
where $\mathcal{A}=\{x,y\}$ and the set of rules $\mathcal{R}$ contains
\[
r_1:\ z \leftarrow x,
\qquad
r_2:\ w \leftarrow z, y,
\]
with $\overline{y}=x$.

The set of ABA arguments includes the following arguments,
written as triples $\alpha=\langle S_\alpha,\mathcal{R}_\alpha,\varphi_\alpha\rangle$:
$\alpha_x := \langle \{x\}, \emptyset, x \rangle$, $\alpha_y := \langle \{y\}, \emptyset, y \rangle$, 
$\alpha_z := \langle \{x\}, \{r_1\}, z \rangle$, 
$\alpha_w := \langle \{x,y\}, \{r_1,r_2\}, w \rangle.
$

According to Definition~19, as illustrated in Fig.3, the SAF induced by this ABA framework is
$\mathcal{F}_2=(A_2,D_2,\mathit{Sub}_2)$,
where:

\begin{itemize}
\item The direct attack relation contains the pair
\[
(\alpha_{x}, \alpha_y) \in D_2,
\]
since $\varphi_{\alpha_{x}}=x=\overline{y}$.

\item The direct subargument relation contains
$(\alpha_x,\alpha_z)$,
$(\alpha_z,\alpha_w),(\alpha_y,\alpha_w)\in \mathit{Sub}_2,
$
reflecting that $\alpha_z$ extends $\alpha_x$ by applying rule $r_1$,
and $\alpha_w$ and $\alpha_y$ extend $\alpha_z$ by applying rule $r_2$.
\end{itemize}

No other direct attacks or direct subargument relations are present.
In particular, $\alpha_{x}$ does not directly attack
$\alpha_z$ or $\alpha_w$.
Nevertheless, since $\alpha_y \in \mathit{Sub}_2^{*}(\alpha_w)$,
the conflict grounded at the atomic argument $\alpha_x$
propagates structurally to $\alpha_w$ via subargument closure.
\end{example}

\begin{figure}[t]
\centering
\begin{tikzpicture}[
  node distance=14mm and 18mm,
  arg/.style={draw, circle, inner sep=1.2pt, minimum size=6.5mm, font=\small},
  attack/.style={->, thick, red, shorten >=2pt, shorten <=2pt},
  sub/.style={->, dashed, thick, blue, shorten >=2pt, shorten <=2pt}
]

\node[arg] (ax) {$\alpha_x$};
\node[arg, right=of ax] (ay) {$\alpha_y$};
\node[arg, below=of ax] (az) {$\alpha_z$};
\node[arg, below=of ay] (aw) {$\alpha_w$};

\draw[attack] (ax) -- (ay);

\draw[sub] (ax) -- (az);
\draw[sub] (az) -- (aw);
\draw[sub] (ay) -- (aw);

\node[draw, rounded corners, align=left, font=\small, anchor=north west]
  at ($(az.south west)+(-0.6,-1.0)$)
  {\textcolor{red}{$\rightarrow$} $D_2$ (direct attack)\\
   \textcolor{blue}{$\dashrightarrow$} $\mathit{Sub}_2$ (direct subargument)};

\end{tikzpicture}
\caption{ABA-induced SAF $\mathcal{F}_2=(A_2,D_2,\mathit{Sub}_2)$.
}
\label{fig:aba-induced-saf-ex3}
\end{figure}

\paragraph{Discussion.}
Examples~\ref{ex:aspic-sub} and~\ref{ex:ABA-SAF} show that both rule-based
and assumption-based formalisms naturally distinguish between
\emph{where a conflict originates} and \emph{which arguments depend on the
conflicted component}.
SAFs abstract precisely this distinction by representing direct attacks
and primitive subargument relations as independent primitives, leaving the
propagation of conflict to be handled semantically rather than
representationally.

\end{document}